\documentclass[sigconf]{acmart}

\usepackage{algorithm}
\usepackage{algorithmic}

\usepackage{multirow}

\AtBeginDocument{%
  }

\setcopyright{acmlicensed}
\copyrightyear{2018}
\acmYear{2018}
\acmDOI{XXXXXXX.XXXXXXX}

\acmConference[Conference acronym 'XX]{Make sure to enter the correct
  conference title from your rights confirmation emai}{June 03--05,
  2018}{Woodstock, NY}
\acmISBN{978-1-4503-XXXX-X/18/06}

\begin{document}
\title{RCLMuFN: Relational Context Learning and Multiplex Fusion Network for Multimodal Sarcasm Detection}



\author{Tongguan Wang}
\affiliation{%
  \institution{College of Informatics, Huazhong Agricultural University}
  \city{Wuhan}
  \country{China}}
\email{wang\_tg@webmail.hzau.edu.cn}

\author{Junkai Li}
\affiliation{%
  \institution{College of Informatics, Huazhong Agricultural University}
  \city{Wuhan}
  \country{China}}

\author{Guixin Su}
\author{Yongcheng Zhang}
\affiliation{%
  \institution{College of Informatics, Huazhong Agricultural University}
  \city{Wuhan}
  \country{China}}

\author{Dongyu Su}
\affiliation{%
  \institution{College of Informatics, Huazhong Agricultural University}
  \city{Wuhan}
  \country{China}}

\author{Yuxue Hu}
\affiliation{%
  \institution{College of Informatics, Huazhong Agricultural University}
  \city{Wuhan}
  \country{China}}

\author{Ying Sha}\authornote{Corresponding Author.}
\affiliation{%
  \institution{College of Informatics, Huazhong Agricultural University}
  \city{Wuhan}
  \country{China}}
\email{shaying@mail.hzau.edu.cn}

\renewcommand{\shortauthors}{Trovato et al.}

\begin{abstract}

Sarcasm typically conveys emotions of contempt or criticism by expressing a meaning that is contrary to the speaker's true intent. Accurate detection of sarcasm aids in identifying and filtering undesirable information on the Internet, thereby reducing malicious defamation and rumor-mongering. Nonetheless, the task of automatic sarcasm detection remains highly challenging for machines, as it critically depends on intricate factors such as relational context. Most existing multimodal sarcasm detection methods focus on introducing graph structures to establish entity relationships between text and images while neglecting to learn the relational context between text and images, which is crucial evidence for understanding the meaning of sarcasm. In addition, the meaning of sarcasm changes with the evolution of different contexts, but existing methods may not be accurate in modeling such dynamic changes, limiting the generalization ability of the models. To address the above issues, we propose a relational context learning and multiplex fusion network (RCLMuFN) for multimodal sarcasm detection. Firstly, we employ four feature extractors to comprehensively extract features from raw text and images, aiming to excavate potential features that may have been previously overlooked. Secondly, we utilize the relational context learning module to learn the contextual information of text and images and capture the dynamic properties through shallow and deep interactions. Finally, we employ a multiplex feature fusion module to enhance the generalization of the model by penetratingly integrating multimodal features derived from various interaction contexts. Extensive experiments on two multimodal sarcasm detection datasets show that our proposed method achieves state-of-the-art performance.

\end{abstract}




\maketitle

\section{Introduction}









\begin{figure}[t]
\centering
\includegraphics[width=1\columnwidth]{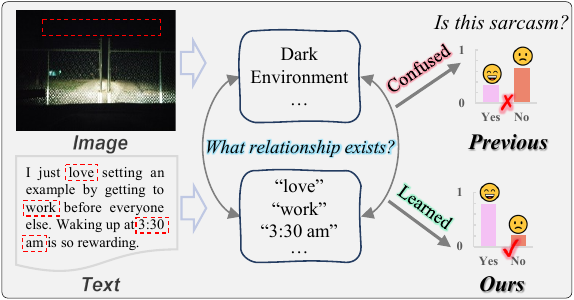} 
\Description{}
\caption{Our method can capture the relational context in the original text (Behaviour of getting up early for work is tightly linked to a love of work, suggesting that getting up at 3:30 am to go to work is not uncommon) and in the image (The black surroundings mean it's still dark).}
\label{fig1}
\end{figure}

Sarcasm is the use of metaphors and hyperbole to reveal or criticize a thing, usually conveying mockery by expressing a meaning opposite to the true intent \cite{1986psycholinguistics, survey}. As shown in Figure \ref{fig1}, the image shows the environment is still dark, but the meaning of the text is “I love waking up at 3:30 am and going to work”, which is an exaggerated way to satirize the unreasonable work schedule. Accurately detecting sarcasm on social media becomes an extremely difficult challenge due to the complex relational context of text and images. Accurately detecting sarcasm content in social media not only improves the performance of downstream tasks, such as sentiment analysis \cite{sentiSuWHZWHS24}, comment moderation \cite{hot}, and human-computer dialogue \cite{robotReimannKOH24} but also is crucial for maintaining the security of the online environment.


Early sarcasm detection methods mainly utilized only textual modal information \cite{silvioAmirWLCS16, SoujanyaPoriaCHV16, MeishanZhangZF16, ChristosBaziotisNPKPEP18}. However, certain groups or individuals are increasingly utilizing multimodal information for sarcasm to intensify the effect of sarcasm, which makes sarcasm detection more challenging. Therefore many researchers have started to use multimodal schemes to accomplish the task of sarcasm detection. The most landmark endeavor is the multimodal sarcasm detection (MMSD) dataset proposed by \citet{cai2019multi}. Based on this dataset, \citet{incrossmgsLiangLL00X21, cmgcnLiangLLY00PX22, dipWenJY23, wei2024g} achieved superior performance by utilizing graph structures to establish entity relationships between text and images. \citet{MMSD2.00001HCCZLCX23} constructed a modified multimodal sarcasm detection dataset (MMSD 2.0) by removing false cues and relabeling irrational samples. At the same time, they proposed a framework called Multi-View CLIP for multimodal sarcasm detection from a fine-grained perspective. \citet{tfcdzhu2024tfcd} adopted a tailored causal graph by generating counterfactual statements and contexts with multimodal sarcasm detection. Leveraging the generative capabilities of LLMs, \citet{LLMSMSD} proposed a multimodal sarcasm model consisting of a designed instruction template to solve the multimodal sarcasm detection task.

Most of the existing research has made extensive use of graph structures with the advantage that such structures can correlate multimodal information into a unified framework. However, real-world scenarios involving text and images are dynamic, with contexts that continually evolve. Meanwhile, the creation of graphs usually relies on external knowledge sources, limiting their effectiveness in recognizing such dynamic updates. Other methods \cite{MMSD2.00001HCCZLCX23, attbertPanL0Q020, cai2019multi,dmsdclJiaXJ24, ICMR} replace the graph structure with some specific strategies for interacting and aligning the extracted multimodal features, but these methods are inadequate for capturing relational contexts (see red box in Figure \ref{fig1}) in the original text and images, which leads to low sarcasm detection accuracy.

To solve above issues, we propose a \textbf{\underline{R}}elational \textbf{\underline{C}}ontext \textbf{\underline{L}}earning and \textbf{\underline{Mu}}ltiplex \textbf{\underline{F}}usion \textbf{\underline{N}}etwork (\textbf{RCLMuFN\footnote{https://github.com/Aichiniuroumian/RCLMuFN.}}) for multimodal sarcasm detection. First, we use four feature extractors separately to fully excavate the potential feature information of the original text and image. Secondly, we use a shallow feature interaction module to interact the image features of the ResNet branch and the text features of the BERT branch respectively. Then, we employ the relational context learning module to capture the dynamic relational context of text and images through both shallow and deep interactions. Finally, we employ a multiplex feature fusion module to enhance the model’s ability to detect sarcasm expressions by penetratingly fusing multimodal features learned from multiple interaction contexts with multimodal features from the CLIP perspective. Experimental evaluation of the MMSD 2.0 dataset demonstrates that our method produces detection accuracy that is 3.91\% higher than the state-of-the-art method, with the same competitive performance on the MMSD dataset.


The main contributions are as follows:

\begin{itemize}

\item We propose a relational context learning and multiplex fusion network for multimodal sarcasm detection (RCLMuFN), which mines, learns, and fuses contextual relationship features between text and images, respectively.


\item We propose a relational context learning module to capture the dynamic relational context of text and images through both shallow and deep interactions.


\item We propose a multiplex feature fusion module to understand sarcasm from various perspectives by penetratingly integrating multimodal features from CLIP-View and containing relational contextual information.

\end{itemize}

\section{Related Work}

\subsection{Unimodal Sarcasm Detection}



Unimodal sarcasm detection methods are divided into two main categories according to the learning process. (1) \textbf{Machine learning methods:} \citet{orenTsurDR10} proposed the SASI algorithm, which combines semi-supervised pattern acquisition and sarcasm classification to perform sarcasm recognition using manually extracted features. \citet{robertoGonzalez-IbanezMW11} concluded that neither machine learning approaches nor human judges perform well on this task by comparing lexical and pragmatic factors. (2) \textbf{Deep learning methods:} \citet{ellenRiloffQSSGH13} improved the recall of sarcasm recognition by automatically guiding the algorithm to learn positive sentiment phrases and negative situation phrases. \citet{mondherBouaziziO15} utilized different components and features of tweets for sarcasm detection. Reduces reliance on complex feature engineering by automatically learning user embeddings and lexical signals. Later \cite{silvioAmirWLCS16, SoujanyaPoriaCHV16, MeishanZhangZF16, ChristosBaziotisNPKPEP18, ChuhanWuWWLYH18, ChuhanWuWWLYH18} all used CNN with the LSTM model to capture long-range dependencies in text for sarcasm detection tasks.

\subsection{Multimodal Sarcasm Detection}


The explosive growth of social media has led to the challenge of detecting sarcasm from posts containing multiple modal messages. \citet{RossanoSchifanellaJTC16} and \citet{SantiagoCastroHPZMP19} demonstrated that the use of multimodal information significantly reduces the sarcasm detection error rate compared to a single modality. \citet{cai2019multi} presented a landmark multimodal sarcasm detection dataset (MMSD) and proposed a hierarchical fusion model for detecting multimodal sarcasm in Twitter. 
Some research by \citet{drnetXuZM20, HongliangPanL0Q020, YuanTianXZM23, dmmdZhuCHLHZ24, milnetQiaoJSCZN23} tried to capture both intra-modal and inter-modal inconsistencies, but still struggle to accurately understand and handle the complex interactions between modalities. To capture the complex relationships within and between modalities, \citet{BinLiangLL00X21, BinLiangLLY00PX22, HKELiuWL22, wei2024g} understood the inter-modal incongruence by introducing graph structure for multimodal sarcasm detection. \citet{dipWenJY23, MMSD2.00001HCCZLCX23, InterCLIP-MEP} exploited sarcasm information in image-text data from multiple perspectives. \citet{dmsdclJiaXJ24} mitigated the effect of text bias through counterfactual data augmentation and adaptive debiased comparative learning mechanism. \citet{tfcdzhu2024tfcd} proposed a TFCD framework to mitigate the effect of dataset bias on multimodal sarcasm detection by using a no-training counterfactual debiasing method. Although the above methods have achieved effective results, the potentially valuable information between text and image is not completely noticed. Meanwhile, the contextual relationship between sarcastic texts and images has not been fully explored, which is a crucial factor in understanding the meaning of sarcasm. Consequently, we propose a relational context learning and multiplex fusion network for multimodal sarcasm detection.

\section{Problem Definition}
We aim to design an effective model for multimodal sarcasm detection, denoted as $\mathcal{M}(*)$, which can accurately recognize if a given sentence and its accompanying image convey sarcasm content. Suppose we have a set of $N$ training samples $\mathcal{D}=\left\{s^{i}\right\}_{i=1}^{N}$, each sample $s^{i}=\left\{{T}^{i}, {V}^{i}, Y^{i}\right\}$ containing three elements. ${T}^{i}$ denotes the textual information, ${V}^{i}$ denotes the visual information, and ${Y}^{i}\in\{0, 1\}$ is the ground truth label of the $i_{th}$ sample, respectively.
\begin{equation}
\mathcal{M}\left({T}^{i}, {V}^{i}\mid\Theta\right)\rightarrow\hat{Y}^{i},
\end{equation}
where $\Theta$ denotes the parameters of the model $\mathcal{M}(*)$, and $\hat{Y}^{i}\in\{0, 1\}$ represents the predicted result, where 0 indicates non-sarcasm and 1 indicates sarcasm.

\section{Methodology}

\begin{figure*}[t] 
\centering 
    \includegraphics[width=1\linewidth]{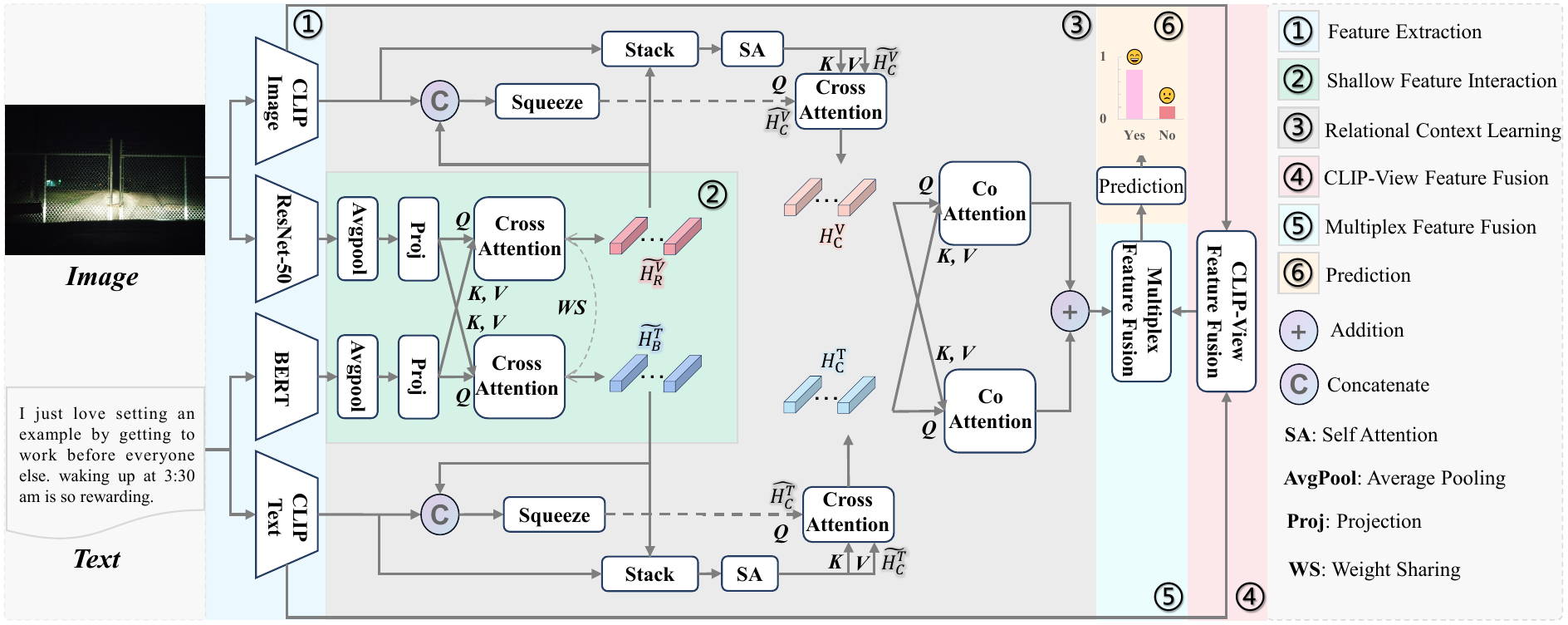} 
    \Description{}
    \caption{Relational Context Learning and Multiplex Fusion Network.}
    \label{fig2} 
\end{figure*}


We propose RCLMuFN to mine potential information for sarcasm detection in the original image-text pairs as much as possible learn the relational context of these pieces of information, and ultimately fuse the features of multiple paths to improve the accuracy of sarcasm detection. The framework of RCLMuFN is shown in Figure \ref{fig2} and consists of six main components: 1) Feature Extraction; 2) Shallow Feature Interaction Module; 3) Relational Context Learning Module; 4) CLIP-View Feature Fusion Module; 5) Multiplex Feature Fusion Module; and 6) Prediction Module.


\begin{algorithm}[h]
\caption{RCLMuFFN Algorithm}
\label{alg:algorithm}
\textbf{Input}: An image-text pair T and V, with its corresponding label Y.
\textbf{Output}: Optimized parameters $\Theta$.
\begin{algorithmic}[1] 

\FOR{each sample} 
\STATE  ${H}_{{C}}^{V}$ is obtained through Eq. (2).
\STATE  ${H}_{{R}}^{V}$ is obtained through Eq. (3).
\STATE  ${H}_{{C}}^{T}$ is obtained through Eq. (4).
\STATE  ${H}_{{B}}^{T}$ is obtained through Eq. (5).

\FOR{${H}_{{R}}^{V}$, ${H}_{{B}}^{T}$}
    \STATE  $\widetilde{H_{R}^{V}}$ is obtained through Eq. (8).
    \STATE  $\widetilde{H_{B}^{T}}$ is obtained through Eq. (9).
\FOR{${H}_{{R}}^{V}$, ${H}_{{B}}^{T}$, $\widetilde{H_{R}^{V}}$, $\widetilde{H_{B}^{T}}$}
\STATE  $\mathcal{H}^{V}$ is obtained through Eq. (12).
\STATE  $\mathcal{H}^{T}$ is obtained through Eq. (15).
\STATE  $\mathcal{H}_{deep}$ is obtained through Eq. (16).
\ENDFOR
\ENDFOR
\FOR{${H}_{{C}}^{V}$, ${H}_{{C}}^{T}$}
\STATE  $\mathcal{H}_{{CLIP}}$ is obtained through Eq. (17).
\ENDFOR
\STATE  $\mathcal{F}_{fuse}^{*}$ is obtained through Eq. (19).
\STATE Compute $\hat{y}$ through Eq. (20).
\STATE Compute $\mathcal{L}_{ce}$ through Eq. (21).
\STATE Apply $\mathcal{L}_{ce}$ to optimize parameters $\Theta$.
\ENDFOR

\end{algorithmic}
\end{algorithm}


\subsection{Feature Extraction}


Given an image-text pair, the text $T=\{w_{1},w_{2},...,w_{n}\}$, where $n$ is the length of the sentence and $V=\{Image\}$. To mine as much potential feature information for detecting sarcasm content from the original text-image pairs. We use two pre-trained feature extractors for raw text and images respectively. For image, we use CLIP-Image encoder \cite{clipradford21a} and ResNet-50 \cite{resnetHeZRS16}, to extract fine-grained semantic and deep features of images, respectively. For text, we use CLIP-Text encoder \cite{clipradford21a} and BERT \cite{bertDevlinCLT19}, to extract broader and more detailed semantic features, respectively. Since CLIP relies on massive image-text pair data with high data diversity, BERT focuses on unimodal language modeling with high data quality for tasks that require in-depth language understanding.
\begin{equation}
{H}_{{C}}^{V} = CLIP(Image),
\end{equation}
\begin{equation}
{H}_{{R}}^{V} = ResNet(Image),
\end{equation}
\begin{equation}
{H}_{{C}}^{T} = CLIP(w_{1},w_{2},...,w_{n}),
\end{equation}
\begin{equation}
{H}_{{B}}^{T} = BERT(w_{1},w_{2},...,w_{n}),
\end{equation}
where $n$ represents the length of the sentence.

\subsection{Shallow Feature Interaction Module}

Since sarcasm typically entails cross-modal implicit meanings and contextual nuances. Consequently, we first employ the pre-trained ResNet-50 \cite{resnetHeZRS16} and the pre-trained BERT \cite{bertDevlinCLT19} to extract raw image features and text features respectively.


As shown in Figure \ref{fig2}, taking the image feature $H_{R}^{V}$ as an example, this pathway feature plays a crucial role in enhancing the representation of the image feature in the relational context learning module in subsequent processes. To maximize the role of this feature, we use the cross-attention module to shallowly fuse this path feature with the text feature $H_{B}^{T}$ extracted by BERT, and obtain the interacted image feature $\widetilde{H_{R}^{V}}$. Specifically, since the original image features obtained using CLIP already contain extensive image features, we first perform an average pooling operation on the image features extracted using the ResNet-50 to reduce the model's dependence on location-specific features and to improve the model's ability to generalize to new sarcasm samples. Then, we map the pooled image feature to the same feature dimension as the text.

\begin{equation}
\widehat{H_{R}^{V}}={Proj}\left({AvgPool}\left(H_{R}^{V}\right)\right),
\end{equation}
\begin{equation}
\widehat{H_{B}^{T}}={Proj}\left({AvgPool}\left(H_{B}^{T}\right)\right),
\end{equation}
where $Proj$ represents the projection operation and $AvgPool$ represents the average pooling operation. 

Subsequently, we use this image feature $\widehat{H_{R}^{V}}$ as the query vector $Q$, and $\widehat{H_{B}^{T}}$ as the key vector $K$ and the value vector $V$. Eventually, we get the image feature $\widetilde{H_{R}^{V}}$ after shallow interaction. Similarly, the text feature ${H}_{{B}}^{T}$ undergoes the same processing as the image feature ${H}_{{R}}^{V}$ to obtain the text feature $\widetilde{H_{B}^{T}}$ after shallow interaction. 
\begin{equation}
\widetilde{H_{R}^{V}}=CA(Q_{\widehat{H_{R}^{V}}},K_{\widehat{H_{B}^{T}}}, V_{\widehat{H_{B}^{T}}}),
\end{equation}
\begin{equation}
\widetilde{H_{B}^{T}}=CA(Q_{\widehat{H_{B}^{T}}},K_{\widehat{H_{R}^{V}}}, V_{\widehat{H_{R}^{V}}}),
\end{equation}
where $CA$ stands for cross-attention module. It is worth mentioning that to capture common features across modalities, we use a weight-sharing strategy when using two cross-attention modules.

\subsection{Relational Context Learning Module}


Multimodal sarcasm detection often involves interpreting the multiple meanings present in images and the contextual nuances of text. Achieving this requires a comprehensive understanding of the semantic relationships within the entire image, as well as the intricate connections between the elements of a complete paragraph of text. At the same time, the model must also capture the dynamic character of the sarcastic meaning by fully grasping the relational context of the visual and textual components.


Therefore, we first learn the context of the relationship between the image and text features internally separately. Then we deeply fuse the image features and text features containing relational contexts through the co-attention module, as a way to mine the deep corresponding meanings between image and text and improve the accuracy of sarcasm detection by deeply understanding the connection between these information. Most importantly, our proposed model captures the dynamically changing nature of sarcasm by learning the relational context of diverse multimodal sarcasm data with better generalization capabilities, as evidenced by the experimental results on the two datasets in Table \ref{table2}.


Specifically, the image features are also used as an example as well. After the previous process, we get the image feature ${H}_{{C}}^{V}$ in CLIP perspective and the image feature $\widetilde{H_{R}^{V}}$ after shallow feature interaction. We fuse these two features by concatenating and stacking operations respectively as follows. 
\begin{equation}
\widehat{H_{C}^{V}}=Squeeze(Concat({H}_{{C}}^{V},\widetilde{H_{R}^{V}})),
\end{equation}
\begin{equation}
\widetilde{H_{C}^{V}}=SA(Stack({H}_{{C}}^{V},\widetilde{H_{R}^{V}})).
\end{equation}

The features after the concatenate operation are obtained by the squeeze operation as $\widehat{H_{C}^{V}}$ as the query vector $Q$. The features after the stack operation are then obtained by the self-attention module as $\widetilde{H_{C}^{V}}$, as the key vector $K$ and the value vector $V$. After the cross-attention module, we get the image feature $\mathcal{H}^{V}$ which contains the rich relationship as shown below.
\begin{equation}
\mathcal{H}^{V}=CA(Q_{\widehat{H_{C}^{V}}}, K_{\widetilde{H_{C}^{V}}}, V_{\widetilde{H_{C}^{V}}}).
\end{equation}

Similarly, the processing flow of text features follows the same approach as that of image features, ultimately resulting in text features $\mathcal{H}^{T}$ that encapsulate the relational context.
\begin{equation}
\widehat{H_{C}^{T}}=Squeeze(Concat({H}_{{C}}^{T},\widetilde{H_{B}^{T}})),
\end{equation}
\begin{equation}
\widetilde{H_{C}^{T}}=SA(Stack({H}_{{C}}^{T},\widetilde{H_{B}^{T}})),
\end{equation}
\begin{equation}
\mathcal{H}^{T}=CA(Q_{\widehat{H_{C}^{T}}}, K_{\widetilde{H_{C}^{T}}}, V_{\widetilde{H_{C}^{T}}}).
\end{equation}

\begin{figure}[t] 
\centering 
    \includegraphics[width=1\linewidth]{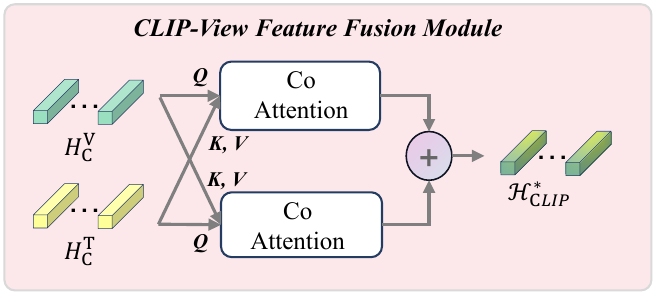} 
    \Description{}
    \caption{CLIP-View Feature Fusion Module.}
        \label{fig3} 
\end{figure}

Finally, to further refine the text and image features that already contain rich relational contexts and enhance their representation of sarcasm, we use two cross-attention modules to compute the interactions between the modalities. This approach allows for a more comprehensive consideration of multimodal information and ultimately improves sarcasm detection performance.
Specifically, we use $\mathcal{H}^{V}$ as the query vector $Q$, and $\mathcal{H}^{T}$ as the key vector $K$ and the value vector $V$. Immediately after that, the processed features are aggregated to form the multimodal feature $\mathcal{H}_{deep}$, which undergoes deep fusion to encapsulate rich relational context.
\begin{equation}
\begin{split}
\mathcal{H}_{deep}=\alpha\cdot CoAttention(Q_{\mathcal{H}^{V}},K_{\mathcal{H}^{T}},V_{\mathcal{H}^{T}})+
\\ (1-\alpha)\cdot CoAttention(Q_{\mathcal{H}^{T}},K_{\mathcal{H}^{V}},V_{\mathcal{H}^{V}}),
\end{split}
\end{equation}
where $\alpha\in[0,1]$ represents the weight of the co-attention module passing through the top of Figure \ref{fig2}.

\subsection{CLIP-View Feature Fusion}



Considering that sarcasm expressions frequently involve the intertwining of linguistic and visual cues, we will first fuse the extracted raw image features $H_{{C}}^{V}$ and textual features $H_{{C}}^{T}$ from the CLIP perspective. The CLIP model has been pre-trained on many image and text data and can extract rich semantic features \cite{clipradford21a}. Notably, when textual cues are insufficient, visual elements often can provide supporting critical insights. Multimodal feature fusion facilitates the detection of sarcasm content while preserving the original hierarchical structure, as illustrated in Figure \ref{fig3}.

Specifically, we use $H_{{C}}^{V}$ as the query vector $Q$, $H_{{C}}^{T}$ as the key vector $K$, and the value vector $V$. We use two co-attention modules to process these sophisticated cues simultaneously. The CLIP-View feature fusion module prioritizes crucial information for sarcasm detection by strategically allocating attention weights within the co-attention module. The processing is shown below. 
\begin{equation}
\begin{split}
\mathcal{H}_{{CLIP}}= \beta\cdot{CoAttention}\left(Q_{H_{{C}}^{V}}, K_{H_{{C}}^{T}}, V_{H_{{C}}^{T}}\right)+\\
(1-\beta)\cdot{CoAttention}\left(Q_{H_{{C}}^{T}},K_{H_{{C}}^{V}}, V_{H_{{C}}^{V}}\right),
\end{split}
\end{equation}
where $\beta\in[0,1]$ represents the weight of the co-attention module passing through the top of Figure \ref{fig3}. The features $\mathcal{H}_{{CLIP}}$ of the fused multimodal CLIP perspectives can capture the semantic links between images and texts at the most original level, which assist the model in discovering the correspondence between the texts and images, and thus improve the accuracy of sarcasm detection.

\subsection{Multiplex Feature Fusion Module}

\begin{figure}[t] 
\centering 
    \includegraphics[width=1\linewidth]{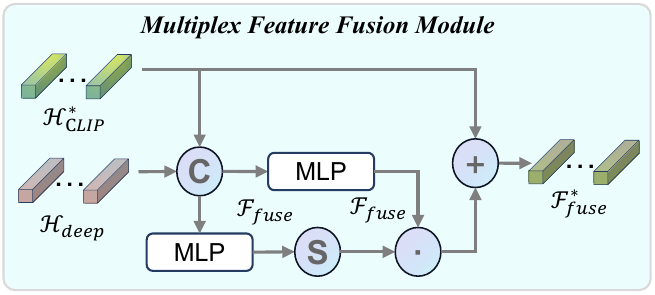} 
    \Description{}
    \caption{Multiplex Feature Fusion Module.}
        \label{fig4} 
\end{figure}

To effectively utilize the extracted and fused multimodal feature information from the previous steps to improve the model's overall comprehension of the sarcasm context, and thus improve the accuracy of detecting sarcasm content, we propose a multiplex feature fusion module to fuse $\mathcal{H}_{{CLIP}}$ and $\mathcal{H}_{deep}$ as shown in Figure \ref{fig4}.


Specifically, after the aforementioned process, we obtain two streams of multimodal features: $\mathcal{H}_{CLIP}$ from the CLIP-View and $\mathcal{H}_{deep}$, which is derived from the relational context learning module and enriched with relational contexts. We first concatenate the two features and use a multilayer perceptron (MLP) layer to process the concatenated features to obtain $\mathcal{F}_{fuse}$ as follows, taking advantage of the fact that MLP can learn complex nonlinear relationships between multimodal features.
\begin{equation}
\mathcal{F}_{fuse}=MLP(Concat(\mathcal{H}_{deep},\mathcal{H}_{{CLIP}})).
\end{equation}

This feature $\mathcal{F}_{fuse}$ is split into two paths, one path goes through the sigmoid function to get a probability, and then multiplied with the other feature $\mathcal{F}_{fuse}$ after going through the MLP layer only, and then finally added with the feature $\mathcal{H}_{{CLIP}}$ to get the final multimodal features used for prediction. Thereby, the model better understands and handles the nuances and complexities in sarcasm as shown below.
\begin{equation}
\mathcal{F}_{fuse}^{*}=\gamma\cdot Sigmoid(\mathcal{F}_{fuse})\cdot\mathcal{F}_{fuse}) + (1-\gamma)\cdot \mathcal{H}_{{CLIP}},
\end{equation}
where $\gamma\in[0,1]$ represents the weight of $\mathcal{H}_{deep}$ this path.


The model can acquire a more comprehensive feature representation by integrating features from diverse sources. This assists in capturing the complexity and diversity of sarcasm content and gaining a more integrated understanding of the nuances between sarcasm and non-sarcasm content. At the same time, it complements the multimodal feature information from different perspectives, thereby enhancing the accuracy of sarcasm detection.

\subsection{Prediction Module}


After passing the fused multimodal features through a fully connected layer, we use the softmax function to convert the predicted score into probability distributions for each sample. Given a text-image pair, the probability of detecting whether or not it contains sarcasm information is calculated from the fused multimodal features $\mathcal{F}_{fuse}^{*}$:
\begin{equation}
\hat{y}=softmax(W\mathcal{F}_{fuse}^{*}+b),
\end{equation}
where $\hat{y}$ is the distribution of output probabilities. $W$ represents the weight of the model, $b$ represents the bias, and $\mathcal{F}_{fuse}^{*}$ represents the final multimodal feature used for prediction.


We use the cross entropy loss function to minimize the cross-entropy between the predicted and true probability distributions to guide the optimization model.
\begin{equation}
\mathcal{L}_{ce} = - y \log(\hat{y}) + (1 - y) \log(1 - \hat{y}),
\end{equation}
where $y\in\{0,1\}$ is the true label of the sample and $\hat{y}=[0,1]$ is predicted probability distribution.


\begin{table}[t]
\caption{Statistics of two experimental datasets.} 
  \label{dataset}
\begin{tabular}{cccc}
    \toprule
\textbf{MMSD/MMSD2.0} & \textbf{Train}         & \textbf{Validation}  & \textbf{Test}        \\ 
    \midrule
Sentences    & 19,816/19,816 & 2,410/2,410 & 2,409/2,409 \\
Positive     & 8,642/9,572   & 959/1,042   & 959/1,037   \\
Negative     & 11,174/10,240 & 1,451/1,368 & 1,450/1,372 \\ 
    \bottomrule
\end{tabular}
\label{table1}
\end{table}

\section{Experiments}

\subsection{Datasets}




To showcase the superiority and generalization of our method, we conduct experiments on two datasets: the MMSD \cite{cai2019multi} and the MMSD 2.0 \cite{MMSD2.00001HCCZLCX23}. The MMSD 2.0 is an upgraded version of the MMSD, having removed misleading factors like hashtags and emoji words. The statistics of these two datasets are presented in Table \ref{table1}.

\section{Experimental Setup}
All experiments were conducted on an NVIDIA Tesla A40 48G with CUDA version 12.2 and pytorch version 2.0.1. The model was implemented based on the huggingface library, using the clip-vit-base-patch32\footnote{https://huggingface.co/openai/clip-vit-base-patch32}. We set both the visual and textual embedding dimensions of CLIP to 768. Additionally, bert-base-uncased\footnote{https://huggingface.co/google-bert/bert-base-uncased} and ResNet-50\footnote{https://huggingface.co/microsoft/resnet-50} were used to encode the text and image features, with the encoded dimensions also set to 768. The image size was uniformly resized to 224×224, and the sequence length of the text encoder was set to 77. During training, we froze the parameters of BERT and ResNet to prevent overfitting. AdamW was used as the optimizer, with a batch size of 32. The model is trained for 10 epochs. The learning rate for the CLIP was set to 1e-6, while for other parts, it was set to 5e-4.

\subsection{Evaluation Metrics}
Based on previous research results, we use Accuracy (Acc), Precision (P), Recall (R), and Micro-average F1 score (F1) to evaluate the model performance. Note that higher values indicate better performance for all metrics.

\begin{table*}[t]
\setlength\tabcolsep{6.8pt} 
\caption{Comparison results (\%) with baseline models on the MMSD and MMSD 2.0 dataset. $\dag$ represents these results cited from the reference \cite{MMSD2.00001HCCZLCX23}. \textsuperscript{*} represents these results cited from their respective references. The best results are highlighted in boldface.} 
\centering
\begin{tabular}{cccccccccc}
\hline
\textbf{\multirow{2}{*}{Modality}}    & \textbf{\multirow{2}{*}{Model}}              & \multicolumn{4}{c}{\textbf{MMSD}}                                          & \multicolumn{4}{c}{\textbf{MMSD 2.0}}                                       \\ \cline{3-10} 
                             &                                     & \textbf{Acc}(\%)        & \textbf{P} (\%)         & \textbf{R} (\%)         & \textbf{F1} (\%)        & \textbf{Acc} (\%)       & \textbf{P} (\%)         & \textbf{R} (\%)         & \textbf{F1} (\%)        \\ \hline
\multirow{3}{*}{Text}        & TextCNN$\dag$\cite{TextcnnKim14}                             & 80.03          & 74.29          & 76.39          & 75.32          & 71.61          & 64.62          & 75.22          & 69.52          \\
                             & Bi-LSTM$\dag$\cite{BilstmGravesS05}                             & 81.90          & 76.66          & 78.42          & 77.53          & 72.48          & 68.02          & 68.08          & 68.05          \\
                             & SMSD$\dag$\cite{SMSDXiongZZY19}                                & 80.90          & 76.46          & 75.18          & 75.82          & 73.56          & 68.45          & 71.55          & 69.97          \\
                              \hline
\multirow{2}{*}{Image}       & ResNet$\dag$\cite{resnetHeZRS16}                              & 64.76          & 54.41          & 70.80          & 61.53          & 65.50          & 61.17          & 54.39          & 57.58          \\
                             & ViT$\dag$\cite{vitDosovitskiyB0WZ21}                                 & 67.83          & 57.93          & 70.07          & 63.40          & 72.02          & 65.26          & 74.83          & 69.72          \\ \hline
\multirow{14}{*}{Multimodal} & HFM$\dag$\cite{cai2019multi} (ACL'19)                                 & 83.44          & 76.57          & 84.15          & 80.18          & 70.57          & 64.84          & 69.05          & 66.88          \\
                             & D\&RNet$\dag$\cite{drnetXuZM20} (ACL'20)                           & 84.02          & 77.97          & 83.42          & 80.60          & -              & -              & -              & -              \\
                             & Att-BERT$\dag$\cite{attbertPanL0Q020} (EMNLP'20)                            & 86.05          & 80.87          & 85.08          & 82.92          & 80.03          & 76.28          & 77.82          & 77.04          \\
                             & InCrossMGs$\dag$\cite{incrossmgsLiangLL00X21} (MM'21)                          & 86.10          & 81.38          & 84.36          & 82.84          & -              & -              & -              & -              \\
                             & CMGCN$\dag$\cite{cmgcnLiangLLY00PX22} (ACL'22)                              & 86.54          & -              & -              & 82.73          & 79.83          & 75.82          & 78.01          & 76.90          \\
                             & HKE$\dag$\cite{HKELiuWL22} (EMNLP'22)                                & 87.36          & 81.84          & 86.48          & 84.09          & 76.50          & 73.48          & 71.07          & 72.25          \\
                             & Multi-view CLIP$\dag$\cite{MMSD2.00001HCCZLCX23} (ACL'23)           & 88.33          & 82.66          & 88.65          & 85.55          & 85.64          & 80.33          & 88.24          & 84.10          \\
                             & DIP\textsuperscript{*}\cite{dipWenJY23}  (CVPR'23)                    & 89.59          & 87.76          & 86.58          & 87.17          & -              & -              & -              & -              \\
                             & MILNet\textsuperscript{*}\cite{milnetQiaoJSCZN23}  (AAAI'23)                  & 89.50          & 85.16          & 89.16          & 87.11          & -              & -              & -              & -              \\
                             & Multi-view CLIP+TFCD\textsuperscript{*}\cite{tfcdzhu2024tfcd} (IJCAI'24) & 89.57          & 84.83          & 89.43          & 88.13          & 86.54          & 82.46          & 87.95          & 84.31          \\
                             & DMSD-CL\textsuperscript{*}\cite{dmsdclJiaXJ24} (AAAI'24)                  & 88.95          & 84.89          & 87.90          & 86.37          & -              & -              & -              & -              \\
                             & G\textsuperscript{2}SAM\textsuperscript{*}\cite{wei2024g} (AAAI'24)                   & 90.48          & \textbf{87.95}          & 89.02          & 88.48          & -              & -              & -              & -              \\
                             & DMMD\textsuperscript{*}\cite{dmmdZhuCHLHZ24} (COLING'24)                   & 90.60          & 86.95          & 91.04          & 88.93          & -              & -              & -              & -              \\ 
                             & Tang et al. \textsuperscript{*}\cite{naaclTangLY024} (NAACL'24)                   & 89.97          & 89.26          & 89.58          & 89.42          & 86.43              & 87.00              &86.30	              & 86.34              \\ 
& CofiPara \textsuperscript{*}\cite{cofipara2024} (ACL'24)                   & -           & -           & -           & -           & 85.70              & 85.96              & 85.55              & 85.89             \\ 
                             
                             \cline{2-10}

                             & \textbf{RCLMuFN (Ours)}                               & \textbf{93.09} & 87.71 & \textbf{95.68} & \textbf{91.52} & \textbf{91.57} & \textbf{89.94} & \textbf{90.55} & \textbf{90.25} \\ \hline
\end{tabular}
\label{table2}
\end{table*}

\subsection{Baseline Models}


We compare three text modality methods, two image modality methods, and thirteen advanced multimodal methods as baselines for comparison. These methods are described in detail below. 



\noindent\textbf{For the text modal method.} We compare the following three methods, \textbf{TextCNN} \cite{TextcnnKim14}, \textbf{Bi-LSTM} \cite{BilstmGravesS05}, \textbf{SMSD} \cite{SMSDXiongZZY19}, where SMSD \cite{SMSDXiongZZY19} used self-matching network to capture the inconsistency information of the sentence by exploring the word-to-word interaction.

\noindent\textbf{For the image modal method.} We compare the following two methods, \textbf{ViT} \cite{vitDosovitskiyB0WZ21} and \textbf{ResNet} \cite{resnetHeZRS16}.

\noindent\textbf{For the multimodal method.} We compare the following thirteen methods. \textbf{HFM} \cite{cai2019multi} utilized image, attribute, and text features extracted by a Bi-LSTM network fused into a single feature vector for prediction. \textbf{D\&RNet} \cite{drnetXuZM20} built a decomposition and relation network to model cross-modal contrasts and semantic associations. \textbf{Att-BERT} \cite{attbertPanL0Q020} designed intermodal attention to capture intermodal inconsistencies. \textbf{InCrossMGs} \cite{incrossmgsLiangLL00X21} determined sentiment inconsistencies within and across modalities by constructing heterogeneous within and across modality maps for each multimodal instance. \textbf{CMGCN} \cite{cmgcnLiangLLY00PX22} built a cross-modal graph convolutional network to understand the incoherence between modalities. \textbf{HKE} \cite{HKELiuWL22} explored atomic-level consistency based on multi-head cross-attention mechanisms and combinatorial-level consistency based on graph neural networks. \textbf{Multi-view CLIP} \cite{MMSD2.00001HCCZLCX23} utilized multi-granularity cues from text, image, and text-image interaction viewpoints for multimodal sarcasm detection. \textbf{DIP} \cite{dipWenJY23} mined sarcasm information on both factual and emotional levels. \textbf{MILNet} \cite{milnetQiaoJSCZN23} utilized the underlying consistency between the two modules to improve performance. \textbf{Multi-view CLIP + TFCD} \cite{tfcdzhu2024tfcd} proposed a training-free counterfactual debiasing framework. \textbf{G\textsuperscript{2}SAM} \cite{wei2024g} utilized global graph-based semantic awareness for sarcasm detection. \textbf{DMSD-CL} \cite{dmsdclJiaXJ24} proposed a debiasing multimodal sarcasm detection framework with contrastive learning aimed at mitigating the deleterious effects of biased textual factors on robust generalization. \textbf{DMMD} \cite{dmmdZhuCHLHZ24} proposed a framework dubbed disentangled multi-grained multimodal distilling for multimodal sarcasm detection. \textbf{CofiPara} \cite{cofipara2024} engaged LMMs to generate competing rationales for coarser-grained pre-training of a small language model on multimodal sarcasm detection. \textbf{\citet{naaclTangLY024}} proposed a generative multimodal sarcasm model consisting of a demonstration retrieval module and a designed instruction template.

\begin{table}[t]
\setlength\tabcolsep{0.6pt} 
\caption{Results of ablation experiments (\%) on two multimodal sarcasm detection datasets. ‘w/o’ is an abbreviation for ‘without’.} 
\begin{tabular}{ccccccccc}
\hline
\multirow{2}{*}{\textbf{Model}} & \multicolumn{4}{c}{\textbf{MMSD}}                                 & \multicolumn{4}{c}{\textbf{MMSD2.0}}                              \\ \cline{2-9} 
                                & \textbf{Acc}   & \textbf{P}     & \textbf{R}     & \textbf{F1}    & \textbf{Acc}   & \textbf{P}     & \textbf{R}     & \textbf{F1}    \\ \hline

w/o RCLM & 88.20 & 79.83 & 93.30 & 86.04 & 88.29 & 87.64 & 84.76 & 86.18 \\
w/o MuFFM  & 91.07 & 85.76 & 92.43 & 88.97 & 89.66  & 87.03  & 89.30 & 88.15  \\
w/o SFIM & 91.28 & 87.40 & 90.70 & 89.02 & 89.50 & 82.67& \textbf{95.66} & 88.69\\
w/o CLIP-VFFM & 75.22 & 62.45  & 91.35  & 74.19 & 60.19  & 71.31  & 53.99  & 45.33   \\
w/o RCLM-Image & 85.97& 82.82& 80.76 & 81.77 & 83.27 & 77.09  & 86.98 & 81.74  \\
w/o RCLM-Text & 86.22 & 81.41 & 83.78 & 82.58 & 88.83  & 89.02 & 84.47 & 86.69 \\
w/o SFIM-Image & 85.33 & 81.53 & 80.65 & 81.09 & 83.40 & 79.63 & 82.55 & 81.06 \\
w/o SFIM-Text & 86.22 & 78.92 & 88.22 & 83.31  & 88.13  & 87.51 & 84.48 & 85.97 \\ \hline
\textbf{RCLMuFN}  & \textbf{93.09} & \textbf{87.71} & \textbf{95.68} & \textbf{91.52} & \textbf{91.57} & \textbf{89.94} & 90.55 & \textbf{90.25} \\ \hline
\end{tabular}
\label{table3}
\end{table}

\subsection{Comparison with Baselines}


To demonstrate the superiority of our proposed RCLMuFN, as shown in Table \ref{table2}, we conduct experiments on two multimodal sarcasm detection datasets. Among the unimodal modal methods, the three text-based modal sarcasm detection methods significantly outperform the two image-based modal sarcasm detection methods in terms of various metrics, which is consistent with the conclusion of previous work, which implies that text is more comprehensible and more informative than images \cite{HKELiuWL22, huunimse}.



In terms of using modal information, all multimodal-based sarcasm detection methods outperform unimodal-based methods. On the MMSD 2.0 dataset, our method improves the accuracy and F1 score by 5.03\% and 3.91\%, respectively, compared to the SOTA method. It demonstrates the superiority of our proposed RCLMuFN. Also on the MMSD dataset, our method improves the accuracy and F1 score compared to the SOTA method by 2.49\% and 2.59\%, respectively. This proves that our methods also have excellent generalization ability. It is worth noting that these methods \cite{incrossmgsLiangLL00X21, cmgcnLiangLLY00PX22, HKELiuWL22, dipWenJY23, milnetQiaoJSCZN23, tfcdzhu2024tfcd, wei2024g, dmmdZhuCHLHZ24} are all using graph structures to establish intra and inter-modal semantic associations and to adapt to new data and relationships by analyzing the relationships of nodes and edges, which in turn enhances to improve the accuracy of multimodal sarcasm detection. However, introducing graph structural models in the network usually requires a complex graph construction process with more computational resources, especially when dealing with large-scale graphs. Unlike these methods, our proposed RCLMuFN uses the relational context learning module proposed in this paper instead of graph structure to learn the relationship between contexts in text and images respectively and finally fuses features from different paths through a multiplex feature fusion module. The comparative experimental results in Table \ref{table2} demonstrate that our approach achieves a more impressive performance without redundant graph construction steps.

\subsection{Ablation Study}

To evaluate the effect of different components, as shown in Table \ref{table3}, we conducted the following ablation experiments. 1) \textbf{w/o RCLM}: We remove the relational context learning module to validate the component's ability to learn relational context. 2) \textbf{w/o MuFFM}: We use the addition operation instead of the multiplex feature fusion module to validate the component's ability to learn features from different perspectives. 3) \textbf{SFIM}: We remove the shallow feature interaction module to measure the contribution of the shallow feature interaction module to the relational context learning process. 4) \textbf{CLIP-VFFM}: We remove the CLIP-View feature fusion module to verify the contribution of multimodal features from the CLIP perspectives for sarcasm detection. 5) \textbf{RCLM-Image \& RCLM-Text}: We remove the image and text features in the relational context learning module to verify the contributions of text and image features, respectively. 6) \textbf{SFIM-Image \& SFIM-Text}: We remove the image and text features in the shallow feature interaction module to verify the contributions of text and image features, respectively.

In general, the complete RCLMuFN framework consistently outperforms its variants, highlighting the importance of synergies between the four modules. The detailed analyses are presented as follows.

\noindent\textbf{For the relational context learning module:} 1) This module enhances the model's ability to capture relationships between contexts, enabling it to better understand the semantic information within the relational contexts of both text and image modalities, leading to improvements in sarcasm detection accuracy; 2) The results from two datasets show that multimodal feature interaction through the back end of the relational context learning module allows the model to deal with dynamic representations that generalize to new multimodal satirical contexts; 3) The relational context information of images plays a more crucial role than that of text.

\noindent\textbf{For the multiplex feature fusion module:} This module provides complementary information about multimodal features from various perspectives by fusing the two features into a more comprehensive multimodal representation, which assists the model in better understanding and detecting sarcasm content.

\noindent\textbf{For the shallow feature interaction module:} 1) The performance improvement of this module on the MMSD 2.0 dataset while still being competitive on the MMSD dataset demonstrates that the module improves the model's ability to generalize to multimodal sarcasm detection; 2) Similar to RCLM, the image features in this module provide more valuable information than the text features.

\noindent\textbf{For the CLIP-View feature fusion module:} 1) The module utilizes the CLIP to understand the semantic consistency between images and text, which assists in detecting sarcasm content that relies on a specific relationship between visual and linguistic elements; 2) The features in this view collaborate with deeply fused multimodal features, which improves the robustness of the model by utilizing complementary information from text and images.





\subsection{Visualization of RCLM}



\begin{figure}[t] 
\centering 
    \includegraphics[width=1\linewidth]{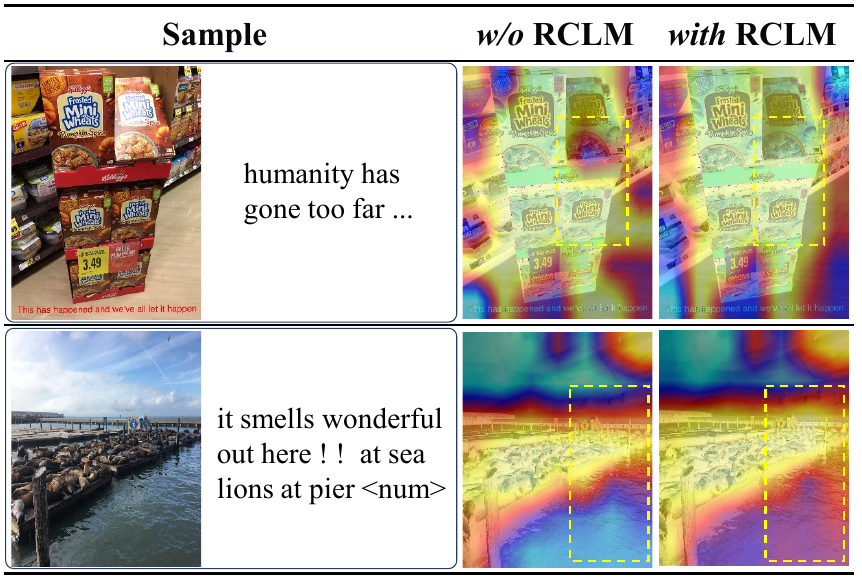} 
    \Description{}
    \caption{Visualization of sample cases using RCLM.}
        \label{fig5} 
\end{figure}

To demonstrate that RCLM enables the model to learn relational contextual information, we visualize the changes in heatmaps before and after applying RCLM for two examples, as shown in Figure \ref{fig5}. In the first example, the spotlight without the RCLM module focuses more on specific local regions, and the image with the RCLM module shows a stronger united focus (see yellow box), which suggests that RCLM pays higher attention to the contextual relationships of the overall image rather than just the local features. RCLMuFN can understand this example satirizes, in a slightly tongue-in-cheek way, the prevalence of unnecessary pumpkin spice in a variety of foods. In the second example, the image shows a lot of sea lions gathered on a pier, and with the addition of RCLM, RCLMuFN can focus on the relationship between the seawater and the seals (see yellow box), which assists the model to understand that this is a satire of the unpleasant odors that may be present in the area.

\subsection{Visualization of MuFFM}



\begin{figure}[t] 
\centering 
    \includegraphics[width=1\linewidth]{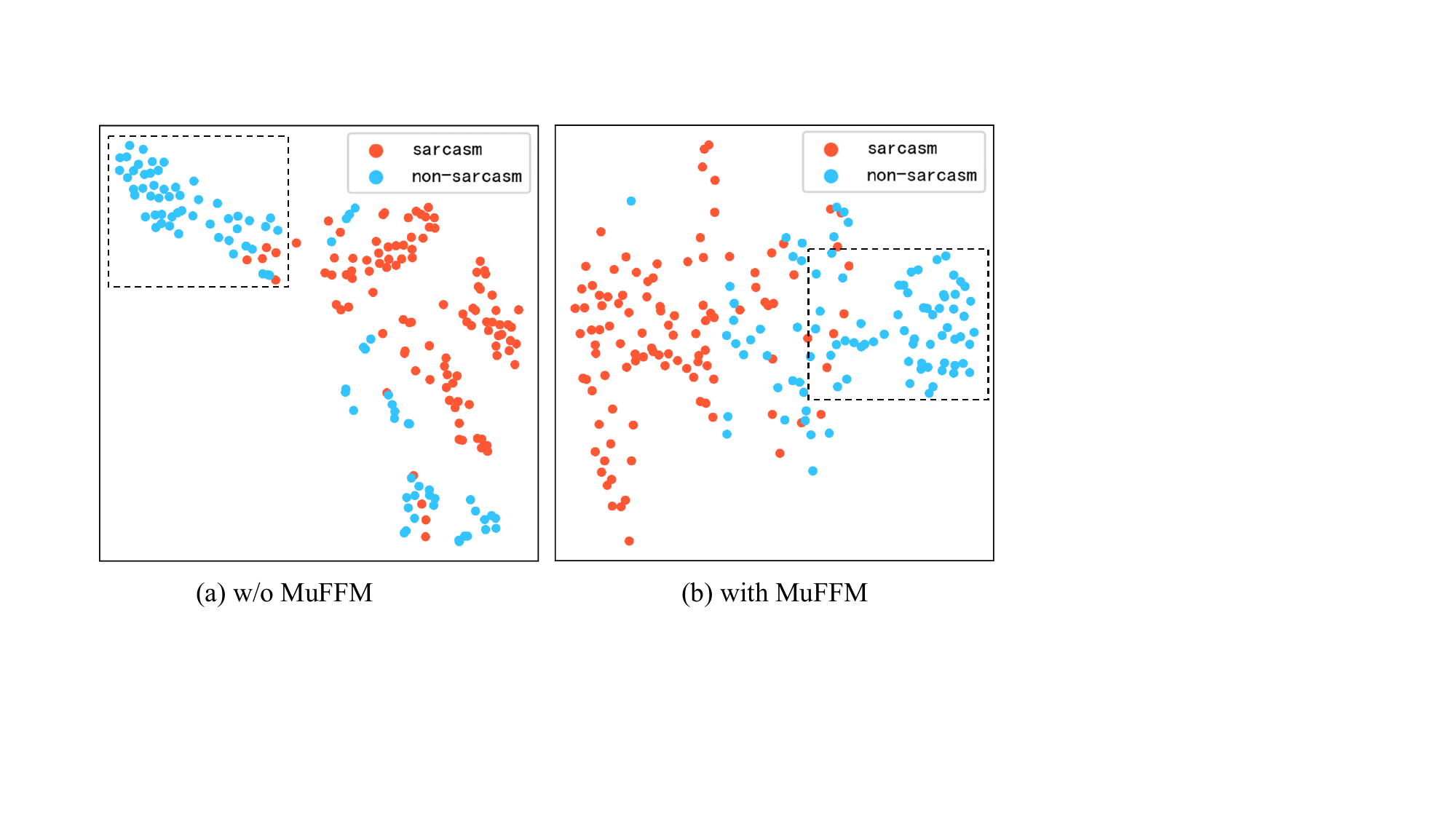} 
    \Description{}
    \caption{Distribution of samples before and after using MuFFM. ‘w/o’ is an abbreviation for ‘without’.}
        \label{fig6} 
\end{figure}

To demonstrate the powerful generalization of MuFFM. As shown in Figure \ref{fig6}, we visualize 200 sarcasm and non-sarcasm samples with MuFFM added and MuFFM removed in the model, respectively. As can be seen from Figure \ref{fig6}(a), if the MuFFM module is removed, although non-sarcasm samples and sarcasm samples can be better distinguished from each other, the feature representations between sarcasm samples and non-sarcasm samples internally are more aggregated, which leads to the model not being well generalized for the detection of a new sample. As can be seen from Figure \ref{fig6}(b), while the non-sarcasm samples can be better distinguished from the sarcasm samples after adding MuFFM, the feature representations between the sarcasm samples and the non-sarcasm samples become more dispersed, which suggests that the model learns a better generalization of the feature representations of the sarcasm samples and the non-sarcasm samples. Thus for a new sample of an unknown content, our model can be more precise in detecting whether it is sarcasm content or not. 



%

\subsection{Error Analysis}

\begin{figure}[t] 
\centering 
    \includegraphics[width=1\linewidth]{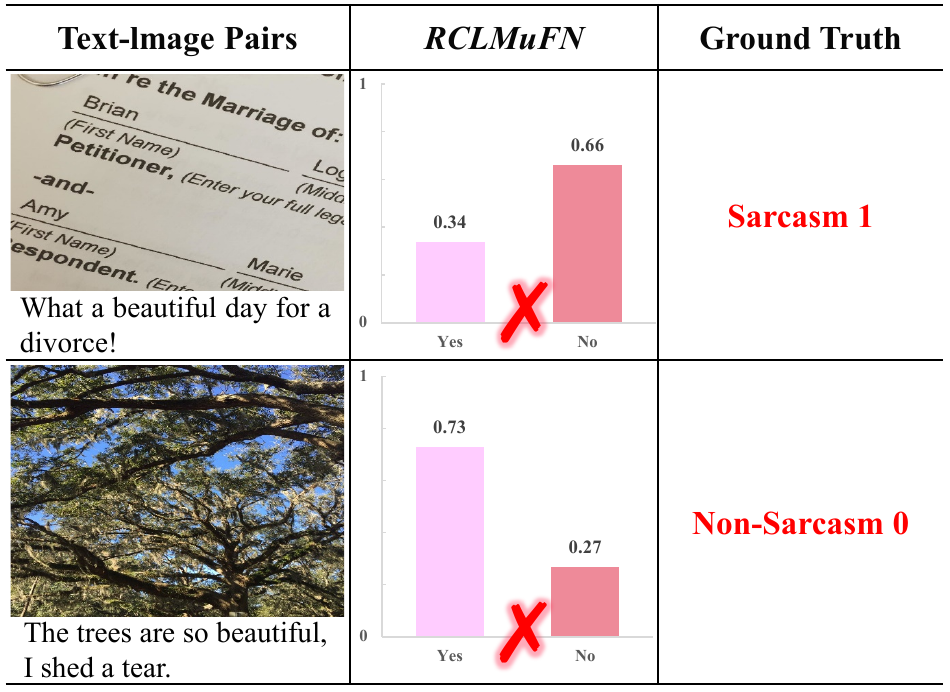} 
    \Description{}
    \caption{Analysis of two error cases.}
        \label{fig9} 
\end{figure}

Although RCLMuFN achieves optimal metrics, there are still challenging sarcasm samples that cannot be detected correctly. We present two samples of failed detection in Figure \ref{fig9}. In the first row, the combined image and text content can be manually evaluated to show that it is used to satirize the event of “divorce” itself, but RCLMuFN detects it as a non-sarcasm sample. In the second line, the combined image and text content is manually evaluated to show that it is only describing the fact that “the tree is beautiful”, which is not sarcasm. We analyze that the common features of these two samples are the short text length and the limited valuable information provided in the image, which poses a challenge to the model in detecting sarcasm. There is still room for improvement in the RCLMuFN in terms of efficiently learning accurate detection of relational context for short textual sarcasm samples.

\subsection{Parameter Analysis}

To assess the impact on model performance when features from different paths are given different weights. We analyzed three weighting coefficients $\alpha$, $\beta$, and $\gamma$ in depth on two sarcasm detection datasets. The performance of the model is evaluated mainly with the Acc and F1 scores as the reference metrics. 

\begin{figure}[t] 
\centering 
    \includegraphics[width=1\linewidth]{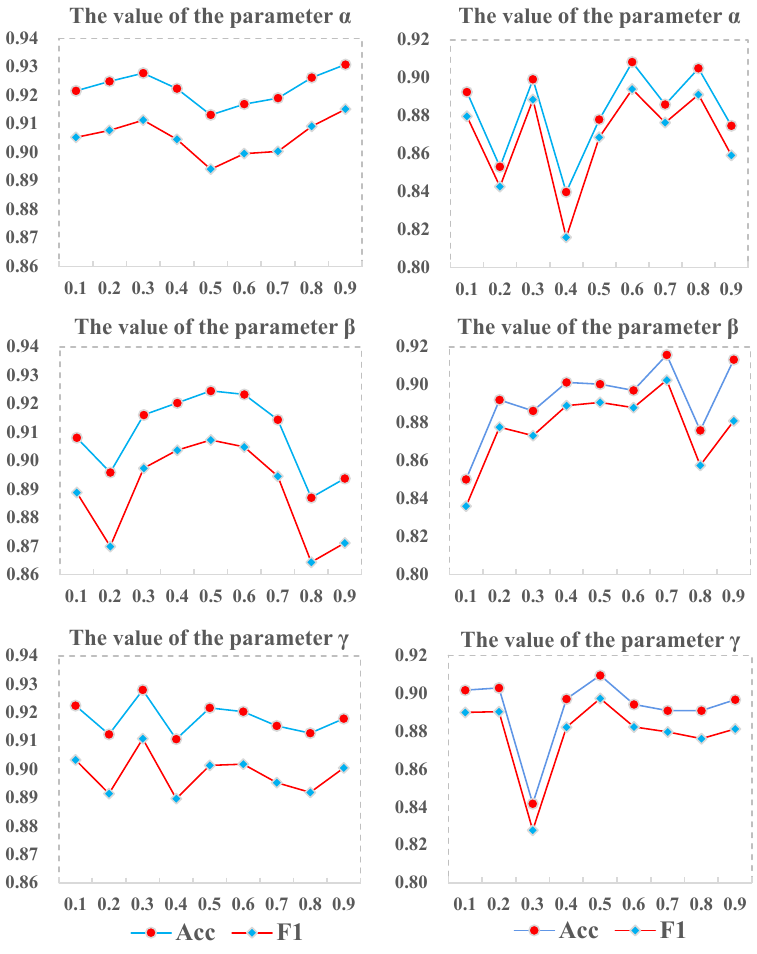} 
    \Description{}
    \caption{ The analysis of the effect of the values of $\alpha$, $\beta$ and $\gamma$ on the performance of the model.}
    \label{fig8}
\end{figure}

\textbf{For} $\alpha$, it controls the weights from the two path features at the end of the RCLM module. As can be seen in the first row of Figure \ref{fig8}, the model performance is optimal when $\beta$ is taken as 0.9 and 0.6 on the MMSD dataset and MMSD 2.0 dataset, respectively. This illustrates the richer relational context information learned by the model from visual features in RCLM. \textbf{For} $\beta$, it controls the weights from the two path features at the end of the CLIP-VFFM. As can be seen from the middle row of Figure \ref{fig8}, the model performance is optimal when $\beta$ is taken as 0.5 and 0.7 on the MMSD dataset and MMSD 2.0 dataset, respectively. It indicates that the information from vision provides more clues for sarcasm detection. \textbf{For} $\gamma$, it controls the weights from the two path features at the end of the MuFFM. As can be seen from the last row of Figure \ref{fig8}, the model performance is optimal when $\beta$ is taken as 0.3 and 0.5 on the MMSD dataset and MMSD 2.0 dataset, respectively. This indicated that the features that passed through RCLM carried a more adequate amount of information in detecting sarcasm content.








\section{Conclusion}


In this paper, we propose a relational context learning and multiplex fusion network (RCLMuFN) for multimodal sarcasm detection. Firstly, we employ four feature extractors to comprehensively extract features from the original text and images, ensuring that excavate potential features that may have been previously overlooked. Secondly, we provide a relational context learning module to capture the relational context of text and images, enabling the model to cope with the dynamic shifts in sarcasm meaning across multiple interactions and enhancing its generalization capability. Finally, we use a multiplex feature fusion module to penetratingly integrate features from text and images after multiple interactions, thereby enhancing the model's ability to detect sarcasm. Extensive experiments on multimodal sarcasm detection datasets demonstrate the effectiveness of RCLMuFN and achieve state-of-the-art performance. We hope that RCLMuFN can provide a novel perspective for future research in other multimodal task studies.

\section{Limitations}


The proposed method in this paper achieves optimal results but still suffers from the following limitations. 1) The method proposed in this paper is currently only for English datasets and may not be effective when dealing with content in other languages. Given the current internationalization trend of social networks, it is necessary to conduct sarcasm detection for various languages in the future. 2) There are still concerns about actual online detection. New sarcasm posts are constantly generated on social media. In the future, it can be considered to combine multimodal large language models to achieve real-time online sarcasm detection.

\section{Ethical Statement}

The MMSD dataset \cite{cai2019multi} and the MMSD 2.0 dataset \cite{MMSD2.00001HCCZLCX23} used in this paper are publicly available datasets, and we respect valuable and creative works in the area of sarcasm detection and other related research. We promise to open-source our code once the paper is accepted. We also expect that our proposed methods will have a positive impact on future related research areas, such as visual understanding, sentiment analysis, and intelligent robots.

\begin{acks}
This work was supported by the National Natural Science Foundation of China (No.62272188), the Fundamental Research Funds for the Central Universities (2662021JC008), and the 2023 Huazhong Agricultural University Independent Science and Technology Innovation Fund Project (2662023XXPY005).
\end{acks}

\bibliographystyle{ACM-Reference-Format}
\bibliography{sample-base}





\end{document}